\documentclass[manuscript,screen]{acmart}
\usepackage{subfigure}
\usepackage{svg}
\AtBeginDocument{%
  \providecommand\BibTeX{{%
    \normalfont B\kern-0.5em{\scshape i\kern-0.25em b}\kern-0.8em\TeX}}}

\setcopyright{acmcopyright}
\copyrightyear{2022}
\acmYear{2022}
\acmDOI{XXXXXXX.XXXXXXX}

\begin{document}

\title{Semi-supervised learning approaches for predicting South African political sentiment
for local government elections}





\author{Mashadi Ledwaba}

\email{u12121879@tuks.co.za}

\affiliation{%
  \institution{Department of Computer Science, University of Pretoria, }
  \country{South Africa}
}

\author{Vukosi Marivate}
\affiliation{%
  \institution{Department of Computer Science, University of Pretoria}
  \country{South Africa}}
\email{vukosi.marivate@cs.up.ac.za}


 \renewcommand{\shortauthors}{M. Ledwaba and V. Marivate}

\begin{abstract}
This study aims to understand the South African political context by analysing the sentiments shared on Twitter during the local government elections. An emphasis on the analysis was placed on understanding the discussions led around four predominant political parties – ANC, DA, EFF and ActionSA. A semi-supervised approach by means of a graph-based technique to label the vast accessible Twitter data for the classification of tweets into negative and positive sentiment was used. The tweets expressing negative sentiment were further analysed through latent topic extraction to uncover hidden topics of concern associated with each of the political parties. Our findings demonstrated that the general sentiment across South African Twitter users is negative towards all four predominant parties with the worst negative sentiment among users projected towards the current ruling party, ANC, relating to concerns cantered around corruption, incompetence and loadshedding.
\end{abstract}




\maketitle

\section{Introduction}
After months of attempts to defer the 2021 planned local government elections in South Africa, the Constitutional Court handed down an order on the 3rd of September 2021 that the elections will take place. Since then, social media platforms such as Twitter have been growing with engagements, analyses, polls, and opinions pertaining to political parties and the elections. With accessible social media data at our disposal, this study aims to understand the South African political context by analyzing public sentiment, particularly when discussing the leading political parties – ANC, DA and EFF. Given the rivalled rise of the newly formed political party, ActionSA, to contend alongside the three giants and with it receiving growing interest from South Africans in a short space of time, it was decided to also include it as the fourth party of the study. 

South Africa has seen insurmountable political unrest in the last couple of years. With the country’s disquieting protests for basic service delivery failures, high unemployment rates and corruption, elections become an important part in ensuring democratic governance and that the right individuals are elected to take on the social and economic challenges that the country is currently facing. This study is important because shared and disseminated sentiment or sentimental language (because it evokes emotion) can influence people’s perceptions, behaviour and voting decisions towards certain political parties. This can help us predict voting intentions and ultimately, parties that stand a good chance of winning the elections. The research questions this study aims to answer regarding South African political sentiment for the elections are as follows:

\begin{itemize}
\item What is the prevalent sentiment towards the four main political parties for the 2021 local government elections?
\item What are the major topics of concern in the negative sentiment tweets towards the political parties?
\item Does sentiment towards a political party drive voting intentions and thus voting results?
\end{itemize}

To achieve the objectives set out above, Natural Language Processing (NLP) was used to extract the public opinions of South
Africans on Twitter towards the political parties to detect negative and positive sentiment. The analysis presents the application of Semi-Supervised Learning (SSL) using graph-based methods to overcome the challenge of using social media data to achieve NLP tasks, where a large amount of the data is unlabelled. This paper presents the analysis in the following structure: a review of related work and methods used in the study, methodology followed, results of the sentiment analysis and topic modelling and lastly, conclusion and limitations of the study.

\section{LITERATURE REVIEW}

Social media platforms such as Twitter, Facebook and Youtube have paved the way for government to interact and develop relationships with citizens. Growth towards e-governance and social networks has been an important driver for public participation, transparent and collaborative governance \cite{mergel_2013}. The adoption of social media in governance has been widely studied for its effectiveness in the dissemination and communication of information as opposed to traditional government websites \cite{yavetz_aharony_2020}, to provide information on opinions of citizens towards government decisions and policies \cite{sharma_shekhar_2020}, its influence on citizens' perception and behaviour \cite{mat_dawi_namazi_hwang_ismail_maresova_krejcar_2021} and its effectiveness during crisis management \cite{guo_liu_wu_zhang_2021}. The focus of this study is on the use of social media to analyse the sentiment of South African citizens in the context of local elections.

\subsection{Sentiment Analysis}
Sentiment analysis is a NLP task of extracting subjective information or opinions expressed in text data – commonly negative and positive sentiment \cite{shanlee_2019}. It is a growing field that has been applied in various industries. Sanders et al. \cite{sanders_2020} used sentiment analysis within the health sector to uncover the public’s attitudes towards the wearing of masks during COVID-19. Mishev et al. \cite{mishev_2020} applied it for text feature extraction to acquire financial signals driven by sentiment and, Bermingham et al. \cite{bermingham-smeaton-2011-using} presents work on monitoring political sentiment to predict elections for the Irish General Elections.

\subsubsection{Sentiment Analysis in Politics}
There has been growing application in the use of various social media platforms to understand and predict behavior within the political context \cite{chambers_2015,franch_2013,jain_kumar_2017}. Franch \cite{franch_2013} demonstrated that the use of tweets in order to infer political sentiment gives a better idea of the political landscape of a country than traditional polls, which often suffer from sample and method bias, and are expensive to conduct than using the freely accessible ‘wisdom of crowd’ offered by social media platforms.

Oyebode et al. \cite{oyebode_orji_2019} presented a comparative study between three lexicon-based classifiers and five machine learning classifiers to determine sentiment in political posts. These were associated with two of the major political parties contending for the Nigerian presidential elections – All Progressives Congress and People’s Democratic Party. For the study, 22 497 posts relating to the presidential candidates of the two
political parties were extracted from their indigenous social media platform (Nairaland). 1041 tweets were randomly chosen from the set and annotated as positive, negative, and neutral to ensure that the annotated set has a balanced representation. VADER and Textblob lexicon-based models were used for
the sentiment classification. The team also addressed the gaps in the coverage of lexicon-based methods by adding 8 748 more features to the VADER lexicon. The obtained results demonstrated that the extended VADER lexicon outperformed the other two (VADER and Textblob) as well as the machine learning approach. 

There are only a few cases of published work centered around South African political sentiment using machine learning approaches. Kotzé et al. \cite{kotze_senekal_2018} presented work on using social media data to extract sentiment and perceptions towards one of South Africa’s Afrikaner minority communities - Orania. Twitter Archiver was used to collect over 10 000 tweets relating to the community. Due to lack of trained data, a lexicon-based approach was followed
for the sentiment analysis. Different sentiment analyzers that employ lexicon dictionaries for sentiment classification were used with the lexicons containing words associated with negative, positive, and neutral sentiment. The lexicon dictionaries that were formulated from actual tweets (such as the NRC word-emotion lexicon and AFINN) outperformed the ones that were extracted from other domains such as the Opinion Lexicon which is based off retail reviews.

The identified gap with the work presented shows that a lexicon-based approach can work well in identifying the polarity of a tweet, however, these publicly available lexicons are not trained on the context of the work studied. Political and social language differs between different social structures and countries therefore, it would be valuable to self-train data on domain-specific text data for the problem. This is typically where semi-supervised learning would be best suited. Given the low-resource gap of sentiment detection tools available where models are pre-trained on political activities in South Africa, this study will be employing semi-supervised methods to intelligently learn the political context relating to the recent local government elections to predict sentiment expressed by South Africans.

\subsubsection{Sentiment Analysis Approach: Semi-Supervised Learning}

There are different methods to automatically detect subjectivity in text data such as pre-trained lexicon-based methods discussed in related work above, machine learning approaches or a hybrid of the two. Even though lexicon-based methods are quick to implement and have been pre-trained on large sets of data, the downside in their use is that they do not cover all domain-specific words. Given that the context of the South African political language is unique and has its own nuances, it would be more suitable to self-train the model on the elections data. 

Most of the online pool of data comes unlabelled, making supervised learning a challenge as it would require labels that represent sentiment. Considering the cost implications in employing human expert annotators and the time constraints for this study, a machine-labelling approach was explored i.e., semi-supervised learning. Semi-supervised learning fits in between supervised and unsupervised learning. It is typically applied when there is a large amount of unlabelled data that is easily accessible than labelled data. The method makes use of the small set of trained labelled data to derive labels on the large amount of the unlabelled data. Zhu et al. \cite{Zhu2002LearningFL} describes semi-supervised classification as having labelled dataset 
\begin{math}
  D_{L}
\end{math} and unlabelled dataset 
\begin{math}
  D_{U}
\end{math} such that you obtain a better performance than you would if you trained a classifier on
\begin{math}
  D_{L}
\end{math}
alone. This approach is particularly suited in this study where there are low resources of pre-trained political sentiment analyzers. The semi-supervised method that will be explored to predict tweet sentiment is a graph-based method known as label propagation.

\subsection{Label Propagation}

Label Propagation (LP) is a transductive learning approach that makes use of known labels in the training process to propagate labels to unlabelled data. LP uses graphs and a small set of labelled nodes in an n-dimensional space connected by edges to find labels for all the nodes in the space given the similarity between the nodes. The algorithm computes a probabilistic transition matrix \textit{T} which gives the probability of a label jumping from node \textit{x} to node \textit{y} i.e., 
\begin{math}
  T_{xy}
\end{math}.
If the probability is high, then \textit{x} and \textit{y} are similar, and the algorithm iteratively propagates labels to training examples by spreading label information through the graph until it achieves global convergence as proposed by Zhu et al. \cite{Zhu2002LearningFL}.

LP has proven to be a robust model in many cases such as Tai et al. \cite{tai_kao_2013} where a sentiment lexicon for sentiment analysis was automatically constructed using LP on unlabelled Twitter data within the financial domain. Experimental results from the study showed that the automatically constructed sentiment lexicon outperformed general-purpose sentiment dictionaries. Other studies from \cite{10.5555/1609067.1609142, yang_shafiq_2018, rajadesingan_liu_2014} also present success in the use of LP for polarity analysis in different domains. 

\section{METHODOLOGY}

This section outlines the experimental set-up aligned to the proposed method to answer the research questions posed. These steps include the data collection process and how it was annotated, data pre-processing and processing steps, sentiment analysis and topic modelling.

\subsection{Data Collection and Annotation}

    Twitter data was collected pre-elections from the beginning of September to end of October 2021. Twitter was chosen in this study due to the ease of extraction of tweets using open-source tools as opposed to Facebook which has more scraping restrictions to protect user data from being leaked. A majority of leaders of the four political parties are more engaged on Twitter, possibly due to their high follower count shown in \cite{batsani-ncube_2021} giving them more reach to citizens. To collect a wider sample of data and counter limitations of each tool, two scraping tools were used - Twint and Twarc. 
    
    Twitter relevant information relating to the four political parties was extracted using various hashtags such as \#MyANC and \#EFFSouthAfrica as well as party leader names (e.g., JSteenhuisen and HermanMashaba). It is important to note that even though ActionSA is not the 4th biggest political party after ANC, DA and the EFF, it was added to the study mainly due to the wide political engagement, social media hype and trends that were present preceding the elections on Twitter relating to the party coupled with it's president being a former member of the DA. Additional data for six other political parties was also collected and the more general tweets relating to the local government elections (using e.g., \#LGE2021).  This additional data was used to enrich our elections corpus for representation however, sentiment was predicted for tweets only associated with the four main political parties. The collected tweets were filtered for non-English tweets and duplicates. From the corpus specific to the four political parties, 1669 tweets were randomly sampled for manual annotation of positive and negative sentiment using, as a guide, an online sentiment lexicon dictionary by Mohammad \cite{mohammad_2021} comprising 54,129 uni-grams of negative and positive sentiment-associated words. Mohammad \cite{mohammad_2016} provides semantic-based questionnaires and techniques also used to guide the annotation process. These annotations were validated by a political commentator and editor who is not affiliated with either of the four political parties but is well-versed in political language and semantics. 
    
    Four ‘different’ datasets were set aside and are given identifiers (A-D) for simplicity of explaining:

\begin{enumerate}
    \item[1.] Dataset A: The complete dataset (labelled and unlabelled) with all the political parties’ election-related data used for data representation purposes.
    \item[2.] Dataset B: Part of Dataset A related only to the four main political parties that will be used for sentiment predictions. Tagging the correct tweet to each political party was very important in this case. To do this, it was ensured that tweets tagged to one political party did not contain any other information about another party or mixed sentiment about two or more political parties using a keyword search and tweet mentions. This method slightly reduces the dataset for the individual parties as was expected given contending parties and conflicting interests where users usually express their opinions towards more than one party in the same tweet. An example tweet of mixed sentiment and different subjects from Dataset B: \textit{"Looks like DA is set to lead SA after the anc. Meanwhile @Julius\_S\_Malema is happy to remain a loud propagandist."}
    \item[3.] Dataset C: Randomly sampled from Dataset B and manually annotated to train the semi-supervised model.
    \item[4.] Dataset D: Randomly sampled from Dataset B and manually annotated and set aside as the hold-out test-set used for model evaluation and acceptance purposes.
 
\end{enumerate}

Table 1 shows a detailed summary of the categorization of datasets.


\begin{table}[htp]
  \caption{Local elections datasets}
\begin{tabular}{@{}clllll@{}}
\toprule
\multicolumn{1}{l}{Dataset} & Purpose                                                                                        & No. of tweets & \begin{tabular}[c]{@{}l@{}}Labelling \\ method\end{tabular} & \begin{tabular}[c]{@{}l@{}}Positive \\ sentiment\end{tabular} & \begin{tabular}[c]{@{}l@{}}Negative \\ sentiment\end{tabular} \\ \midrule
A                           & \begin{tabular}[c]{@{}l@{}}Processing elections data\\  - all political parties\end{tabular}              & 681,029 & Automatic & \multicolumn{2}{c}{N/A} \\
B                           & \begin{tabular}[c]{@{}l@{}}Sentiment predictions for\\ the four political parties\end{tabular} & 401,916 & Automatic & \multicolumn{2}{c}{N/A} \\
C                           & \begin{tabular}[c]{@{}l@{}}Label propagation \\ model training\end{tabular}                    & 1669 & Manual & 720 & 949 \\
D                           & \begin{tabular}[c]{@{}l@{}}Label propagation \\ model testing\end{tabular}                     & 372 & Manual &  181 & 191 \\ 
\bottomrule
\end{tabular}
\end{table}

\subsection{Pre-processing}

A primary challenge associated with NLP tasks relates to handling out-of-range or non-standard text data. Social media text makes it even more of a challenge as out-of-range data comes in many forms – emojis, punctuation, misspelling, URL’s, colloquial language, and other non-standard forms of texting. To handle these, various pre-processing methods were applied to make the text data suitable for analysis and predictability. These included the removal of unwanted texts and symbols in the tweet such as usernames, URL’s, hashtags, numbers, and punctuation. The text was standardized by making it all lower case, contractions where expanded, ticks and the successive letters removed, and extra white spaces removed. Successive words like ‘action’ and ‘sa’ were joined to not lose the representation of ActionSA in some tweets. Stop-words were also removed from the text. These are frequent low-level information words that form part of the language syntax but do not add any semantic meaning to the text such as ‘the’, ‘a’, ‘because’ or ‘have’. Lastly, the text was tokenized and lemmatized. The former is to reduce inflection in words with the same meaning being used in different forms due to grammatical correctness of sentences by representing them in their common root form e.g., the words; \textit{‘corrupt’, ‘corrupts’, ‘corrupted’, ‘corrupting’ \emph{stem from the one word} ‘corrupt’}.

\subsection{Data Processing – Word Embeddings}
Word embeddings are used to create numerical features i.e., vector representation of the text data to transform the text into a machine-readable format. Two word-embedding models that were explored for the representation of the elections data were TF-IDF Encoding and Word2Vec Embeddings.

\subsubsection{TF-IDF}

Term Frequency Inverse Document Frequency (TF-IDF) is an information retrieval technique that measures the relevance of a word in a corpus based on the number of times it appears. TF-IDF determines the relative frequency of words in a document compared to the inverse proportion of those words across the entire corpus of documents \cite{qaiser_ali_2018}. TF-IDF aims to reduce the weight of words that occur frequently in the corpus that do not provide relevant information about the document such as stop-words. For a term \textit{t} in document \textit{d}, the TF-IDF weight \begin{math}
    W(t,d)
    \end{math}, is given as:
\begin{displaymath}
  W(t,d) = TF(t,d) * IDF(t)
\end{displaymath}

where:

\begin{itemize}
    \item Term frequency:    
    \begin{math}
    TF(t,d)
    \end{math} is the number of occurrences of the term \textit{t} in document \textit{d}.
    \item Inverse document frequency: 
    \begin{math}
  IDF(t) = \log(\frac{n}{df(t)})
    \end{math} measures how significant the term is in the corpus where \textit{n} is the total number of documents in the corpus and     
    \begin{math}
    df(t)
    \end{math} 
    is the document frequency of term \textit{t}.
\end{itemize}

\subsubsection{Word2Vec Embeddings}

Word2Vec representations are computed from prediction-based models. Word2Vec was pre-trained on over 100 billion words from the Google News dataset and uses a shallow two-layer neural network to derive vector representations \cite{google_word2vec}. The difference between Word2Vec and TF-IDF is that Word2Vec measures the semantic and syntactic similarities between documents which considers the context that words are used in when representing them therefore, words that share the same context will have similar vector representations. Word2Vec uses two different methods to create the vector representations – Continuous Bag of Words (CBOW) and Skip-gram. CBOW tries to predict the next word in a sentence by considering the context that the target word is being used in as the input into the network. Skip-gram uses the target word to predict the context and produce the representation. Skip-gram takes in as input the one-hot-encoded vector input of the word and gives as an output the probability score of the word being used in the same context as the output layer. CBOW does the opposite by taking in the one-hot-encoded context words to give the probability score of the output word being in the center of the context.

To represent the elections data in vector form, both TF-IDF and Word2Vec (Skip-gram and CBOW) models were fit to Dataset A (the complete dataset across all political parties) allowing for the models to be trained for sentiment predictions.


\subsection{Modelling}

\subsubsection{Supervised Learning}

To develop a baseline to benchmark the Label Propagation model against, a Support Vector Machine (SVM) model is trained. The SVM model is trained on Dataset C (the labelled training set) using the word embeddings and validated using a 5-fold cross validation method. The models are further tested on the hold-out set from Dataset D and the performance yielded by each of these methods was compared for model selection purposes. Metrics used for the evaluation of the models were precision, recall and F1 score \cite{stapor_2017}.

\subsubsection{Semi-Supervised Learning}

The semi-supervised learning method takes place in the following steps:

\begin{enumerate}
    \item[1.] Un-labelling 50\% of the manually labelled data. These training examples unlabelled were randomly sampled from Dataset C.
    \item[2.] Training the semi-supervised model on a combination of the remaining 50\% of labelled training examples and the above-mentioned unlabelled training examples i.e., training the model on Dataset C where 50\% of the data is unlabelled.
    \item[3.] Propagation of sentiment labels to the unlabelled examples.
    \item[4.] Evaluating the transduction of the actual labels that were removed in step 1.
    \item[5.] Evaluation of the semi-supervised model against the baseline supervised models. \item[6.] Testing model performance against the hold-out set.
    \item[7.] Using the semi-supervised model to label an additional batch of randomly sampled data from the unlabelled Dataset B and retraining the model and using it to label another batch, iteratively testing it against the hold-out set until Dataset B is completely labelled.
\end{enumerate}
 
The modelling process of the semi-supervised learning is summarized in the following flow diagram:

\begin{figure}[h]
  \centering
  \includegraphics[width=\linewidth]{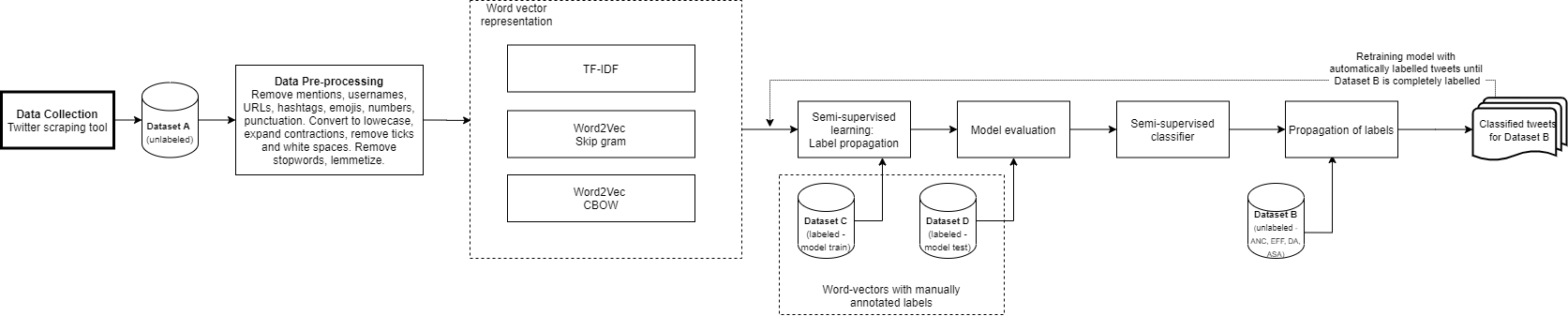}
  \caption{Methodology flow diagram for SSL}
\end{figure}

\subsubsection{Topic Modelling}
The second research question involves uncovering the topics that drive negative sentiment to enable us to better understand the public’s main concerns with each of the political parties. The method used to find the hidden topics in the tweets associated with negative sentiment was Latent Dirichlet Allocation (LDA). LDA is a commonly used unsupervised method for latent topic extraction. LDA views documents as a probabilistic distribution of a set of topics and each topic as a distribution of words. Wang et al. \cite{wang_mccallum_wei_2007} proposed a model on topical n-grams which generates uni-grams and bi-grams based on the context of a document e.g., the model could get the correct context of the bi-gram ‘white house’ as an association to politics rather than real-estate. Within the South African political context, LDA was used between two national election cycles (2014 and 2019) to uncover the predominant recurring topics and themes in news articles in the different election periods  \cite{moodley_marivate_2019}.

\section{Results}
\subsection{Word-Embedding}

Table 2 is a comparison of the topmost cosine similarities to the name of the political party from the Word2Vec
embeddings - CBOW and Skip-gram. It is observed that Skip-gram produced more contextual words commonly associated with each of the political parties than CBOW.

\begin{table}[htp]
\caption{Cosine similarity comparisons between CBOW and Skip-gram Word2Vec}

\begin{tabular}{@{}cll@{}}
\toprule
\multicolumn{1}{l}{Search word} & CBOW                                                                                     & Skip-gram                                                                                                                  \\ \midrule
‘anc’                           & \textit{\begin{tabular}[c]{@{}l@{}}myanc, cancer, corrupt, crook, ret\end{tabular}}   & \textit{\begin{tabular}[c]{@{}l@{}}myanc, ancorruption, ancterriosim, \\ ancriminals, lamasela\end{tabular}}               \\
‘da’                            & \textit{\begin{tabular}[c]{@{}l@{}}ff, vf, nmb, coalition, ct\end{tabular}}           & \textit{\begin{tabular}[c]{@{}l@{}}ff, ffplus, vfplus, midvaal, \\ cassim\end{tabular}}                                    \\
‘eff’                           & \textit{\begin{tabular}[c]{@{}l@{}}actionsa, da, malema, mighty, waya\end{tabular}}   & \textit{\begin{tabular}[c]{@{}l@{}}effmustrise, groundforces, effwayawaya, \\ stockvel, flopper\end{tabular}}              \\
‘actionsa’                      & \textit{\begin{tabular}[c]{@{}l@{}}action4sa, asa, herman, split, party\end{tabular}} & \textit{\begin{tabular}[c]{@{}l@{}}action4sa, itstimeforaction, voteactionsa, \\ mashaba, letsfixsouthafrica\end{tabular}} \\ \bottomrule
\end{tabular}
\end{table}

\subsection{Modelling}
Table 3 shows the mean of cross-validation F1 scores using 5-folds for the baseline and semi-supervised models trained using TF-IDF and Word2Vec representations. The validation scores shows that the models generalise well, with Word2Vec trained models yielding better performance in most cases than TF-IDF models.

\begin{table}[htp]
\caption{Model validation results}
\begin{tabular}{@{}lc@{}}
\toprule
\multicolumn{1}{c}{Model} & \multicolumn{1}{l}{Validation score} \\ \midrule
TF-IDF (SL) - baseline    & 0.85                                 \\
TF-IDF (SSL)              & 0.82                                 \\
SKIPGRAM (SL) - baseline  & 0.89                                 \\
SKIPGRAM (SSL)            & 0.90                                 \\
CBOW (SL) - baseline      & 0.85                                 \\
CBOW (SSL)                & 0.87                                 \\
\bottomrule
\end{tabular}
\end{table}

Table 4 shows a comparative breakdown of the results of the tuned classification models by sentiment polarity. The models for sentiment predictions generally perform well against the independent hold-out set. This could be attributed to having a good vector representation providing rich context of the elections. Cases where the baseline model outperformed the semi-supervised model in precision resulted in significantly unbalanced prediction errors between the false negative and false positive sentiment predictions. The Skip-gram semi-supervised model shows good performing results in most of the metrics categories with more balanced prediction errors on the hold-out set and was used going forward to label the rest of the unlabelled Dataset B (dataset for each of the main political parties).

\begin{table}[]
\caption{Model performance against hold-out set by sentiment polarity}
\begin{tabular}{@{}lcccccc@{}}
\toprule
\multicolumn{1}{c}{Model} & \multicolumn{2}{c}{Precision}                               & \multicolumn{2}{c}{Recall}                                  & \multicolumn{2}{c}{F1 Score}                                \\ \midrule
                          & \multicolumn{1}{l}{Positive} & \multicolumn{1}{l}{Negative} & \multicolumn{1}{l}{Positive} & \multicolumn{1}{l}{Negative} & \multicolumn{1}{l}{Positive} & \multicolumn{1}{l}{Negative} \\ \midrule
TF-IDF (SL) - baseline    & 0.93 & 0.89 & 0.87 & \textbf{0.94} & 0.90 & 0.91\\
TF-IDF (SSL)              & 0.88 & 0.93 & \textbf{0.92} & 0.90 & 0.90 & 0.92\\
Skipgram (SL) - baseline  & \textbf{0.95} & 0.89 & 0.87 & 0.91 & \textbf{0.91} & 0.92\\
Skipgram (SSL)            & 0.89 & \textbf{0.94} & \textbf{0.92} & 0.91 & \textbf{0.91} & \textbf{0.93} \\
CBOW (SL) - baseline      & \textbf{0.95} & 0.88 & 0.87 & \textbf{0.94} & 0.90 & 0.91 \\
CBOW (SSL)                & 0.94 & 0.89 & 0.88 & \textbf{0.94} & \textbf{0.91} & \textbf{0.93} \\ 
\bottomrule
\end{tabular}
\end{table}

\subsubsection{Automatic Labelling of Data}
Three labelling iterations were conducted using the best performing semi-supervised model with the Skip-gram representation. The first iteration used randomly selected 1000 training examples from the unlabelled Dataset B for each of the four political parties. Ideally, we want the model to generalise well across the different parties therefore, the random samples came from the four respective sub-datasets of each political party i.e., 250 data points sampled from each political party. The model is used to propagate labels onto the 1000 examples and then retrained on a combined set of the newly labelled 1000 examples and the initial training set, and tested against the hold-out set to ensure acceptable predictive performance. This procedure was followed to label an additional 10 000, followed by 20 000 random training examples from each political party’s respective datasets. 

\subsubsection{Sentiment Analysis Results}
The final retrained model after the third labelling and retraining iteration showed acceptable results against the independent hold-out set:

\begin{table}[htp]
\caption{Confusion matrix of final SSL model against hold-out set}
\begin{tabular}{@{}llll@{}}
\toprule
                  & Positive (prediction) & Negative (prediction) & All \\ \midrule
Positive (actual) & 150                   & 31                    & 181 \\
Negative (actual) & 14                    & 177                   & 191 \\
All               & 164                   & 208                   & 372 \\ 
\bottomrule
\end{tabular}
\end{table}

The remaining examples of the datasets for the four political parties were automatically labelled using the semi-supervised model. From the obtained sentiment results (Table 6), it is observed that the general sentiment of the South African political context for the local government election relating to the parties is negative. ANC is the current ruling party and is the most discussed party generally generating more tweet data than the other contenders therefore, the results will be interpreted in terms of the sentiment relative to the number of tweets generated for the party (i.e., sentiment percentage). Tweets associated with the ANC show the worst negative sentiment percentage compared to all the parties which highlights the level of dissatisfaction with the ruling party. The newly formed party, ActionSA, has a more positive tweet percentage compared to the other parties.

\begin{table}[htp]
\caption{Sentiment classification for the four political parties}
\begin{tabular}{@{}clll@{}}
\toprule
Political party & Total labelled & Positive sentiment & Negative sentiment \\ \midrule
ANC             & 195,342        & 51,407 (26\%)      & 143,935 (74\%)     \\
EFF             & 87,430         & 33,825 (39\%)      & 53,605 (61\%)      \\
ActionSA        & 60,990         & 28,756 (47\%)      & 32,234 (53\%)      \\
DA              & 58,154         & 21,545 (37\%)      & 36,609 (63\%)      \\ \midrule
Total           & 401,916        & 135,533 (34\%)     & 266,383 (66\%)     \\ \bottomrule
\end{tabular}
\end{table}

\begin{figure}[htp]
  \centering
  \includegraphics[width=0.5\linewidth]{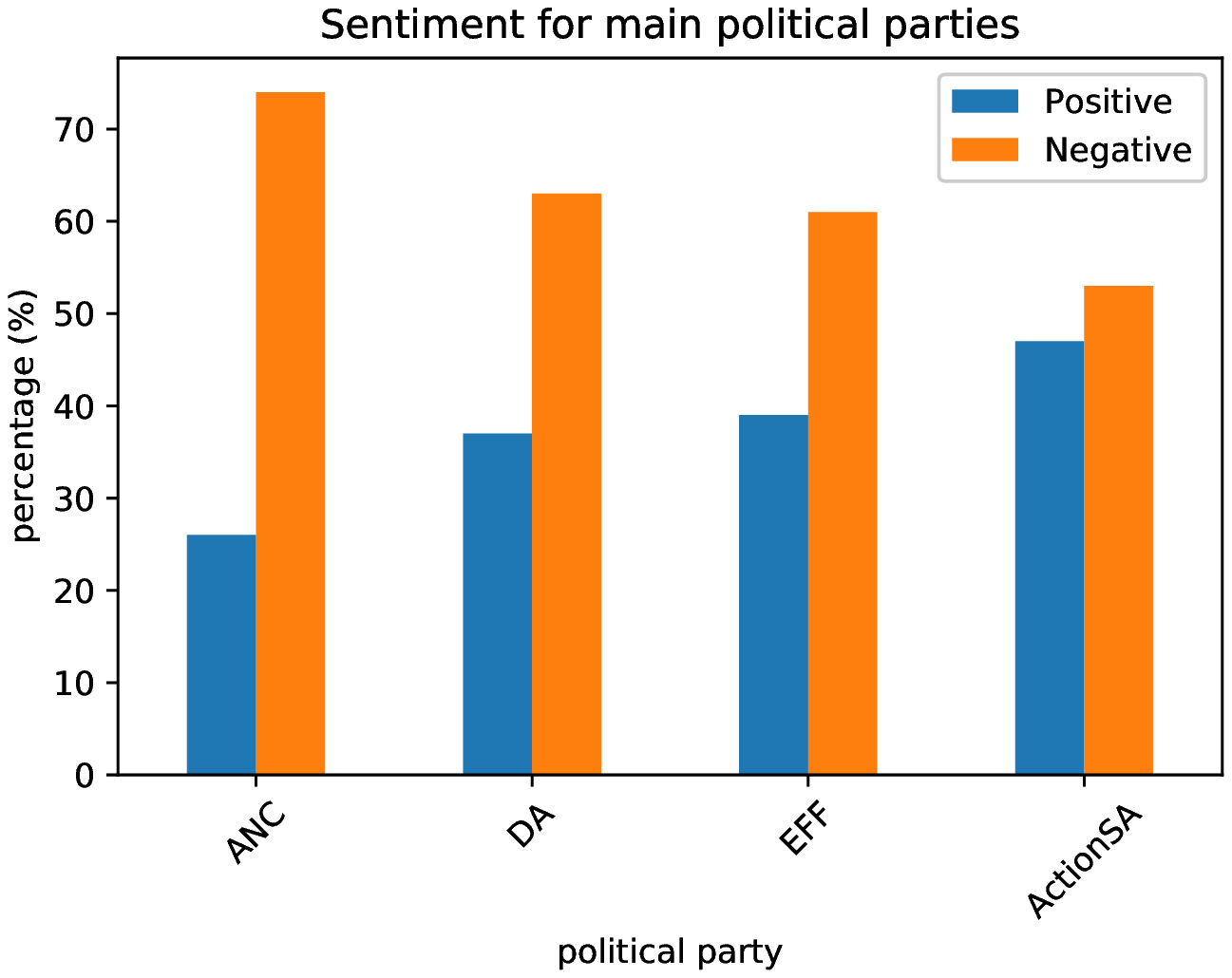}
  \caption{2021 local government election sentiment results}
\end{figure}

\subsubsection{Topic Modelling}
The task of uncovering topics associated with the negative sentiment expressed towards each of the four political parties was done using the initial TF-IDF model. The results shown and discussed for each political party include a word cloud showing the importance of the words in the context or topics. Another method to present the recurring topics for the political parties is to extract the frequently occurring sequence of words (n-grams). To get more context, 4-grams were extracted.

\paragraph{ANC}
The negative sentiment projected towards the ANC is associated with themes centered around corruption, incompetence and loadshedding, where the latter was a predominant theme uncovered in the 2019 national elections as well \cite{moodley_marivate_2019}. The extracted top 4-grams show the extent of the public's dissatisfaction with ANC where corruption is a major theme across most of the top 4-grams. Other themes in the topic model covered unemployment concerns, Covid and lockdowns.

\begin{figure}[htp]
    \centering
    \subfigure[]{\includegraphics[width=0.2\linewidth, height = 3.5cm]{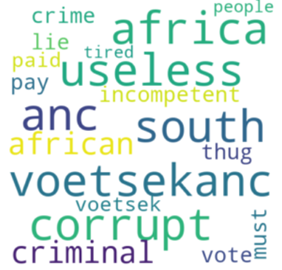}}
    \subfigure[]{\includegraphics[width=0.2\linewidth, height = 3.5cm]{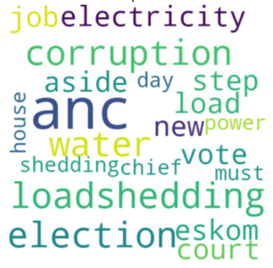}}
    \subfigure[]{\includegraphics[width=0.5\linewidth, height = 3.5cm]{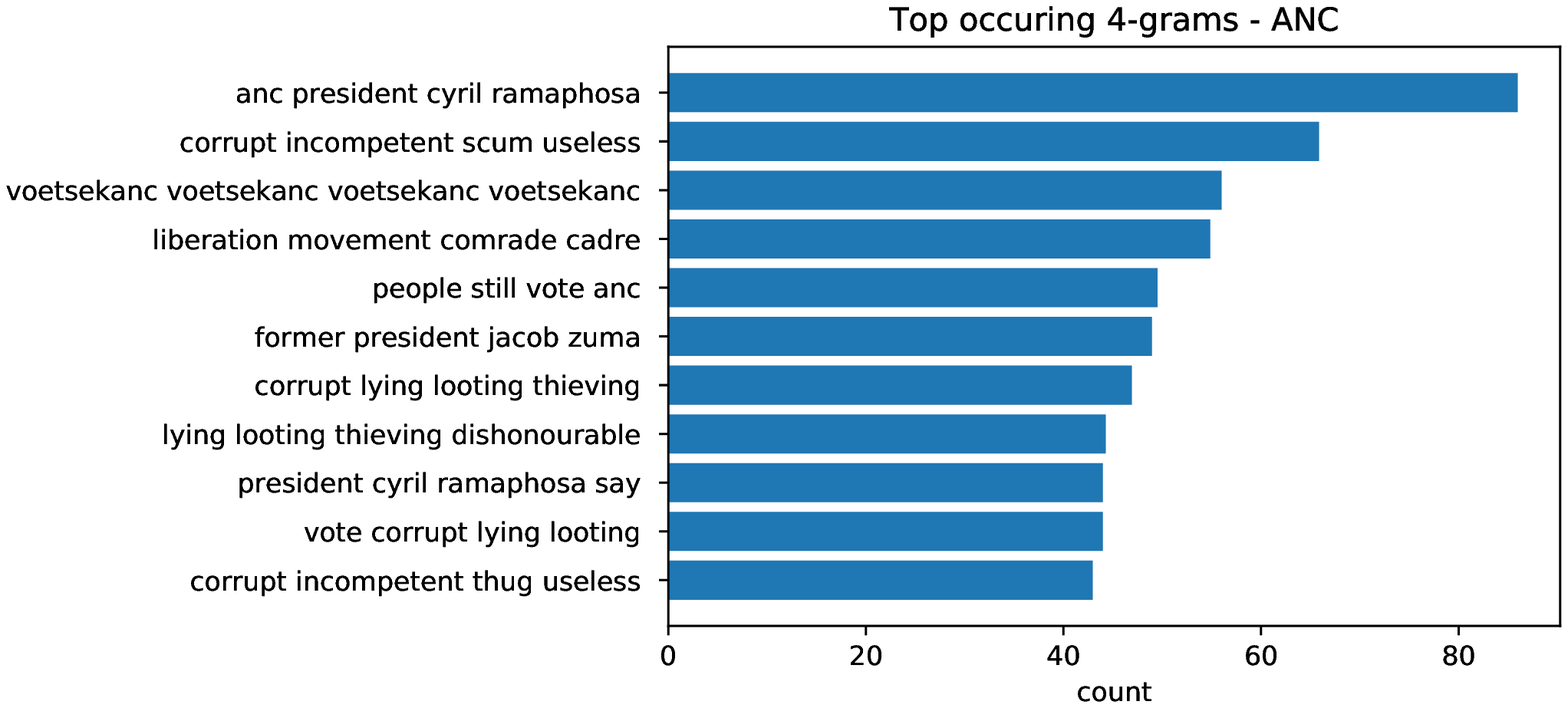}}
    \caption{ANC negative sentiment topics word cloud (a),(b) and ANC recurring themes using 4-grams (c)}
\end{figure}

\paragraph{DA} Negative sentiment tweets for the DA include the names of Gareth Cliff and John Steenhuisen as observed in the 4-grams for the party. Those negative tweets are likely to be linked to an interview a week before elections between Gareth Cliff, Johan Steenhuisen and community activist Mudzuli Rakhivhane where Gareth Cliff made a statement insinuating that racism in South Africa is not as big of a problem as service delivery failures. Cliff’s dismissive notion resulted in a social media outcry from the South African public with most of the outrage centered on racism – a recurring theme associated with the DA. The theme of racism is also associated with an election poster that was erected in Phoenix referencing the events of the July unrest. The poster caused major social media uproar branding DA as a racist party
for calling community members involved in the violence that unfolded as heroes.

\begin{figure}[htp]
    \centering
    \subfigure[]{\includegraphics[width=0.2\linewidth, height = 3.5cm]{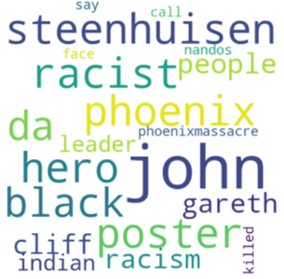}}
    \subfigure[]{\includegraphics[width=0.2\linewidth, height = 3.5cm]{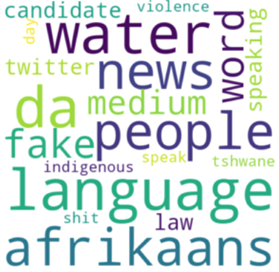}}
    \subfigure[]{\includegraphics[width=0.5\linewidth, height = 3.5cm]{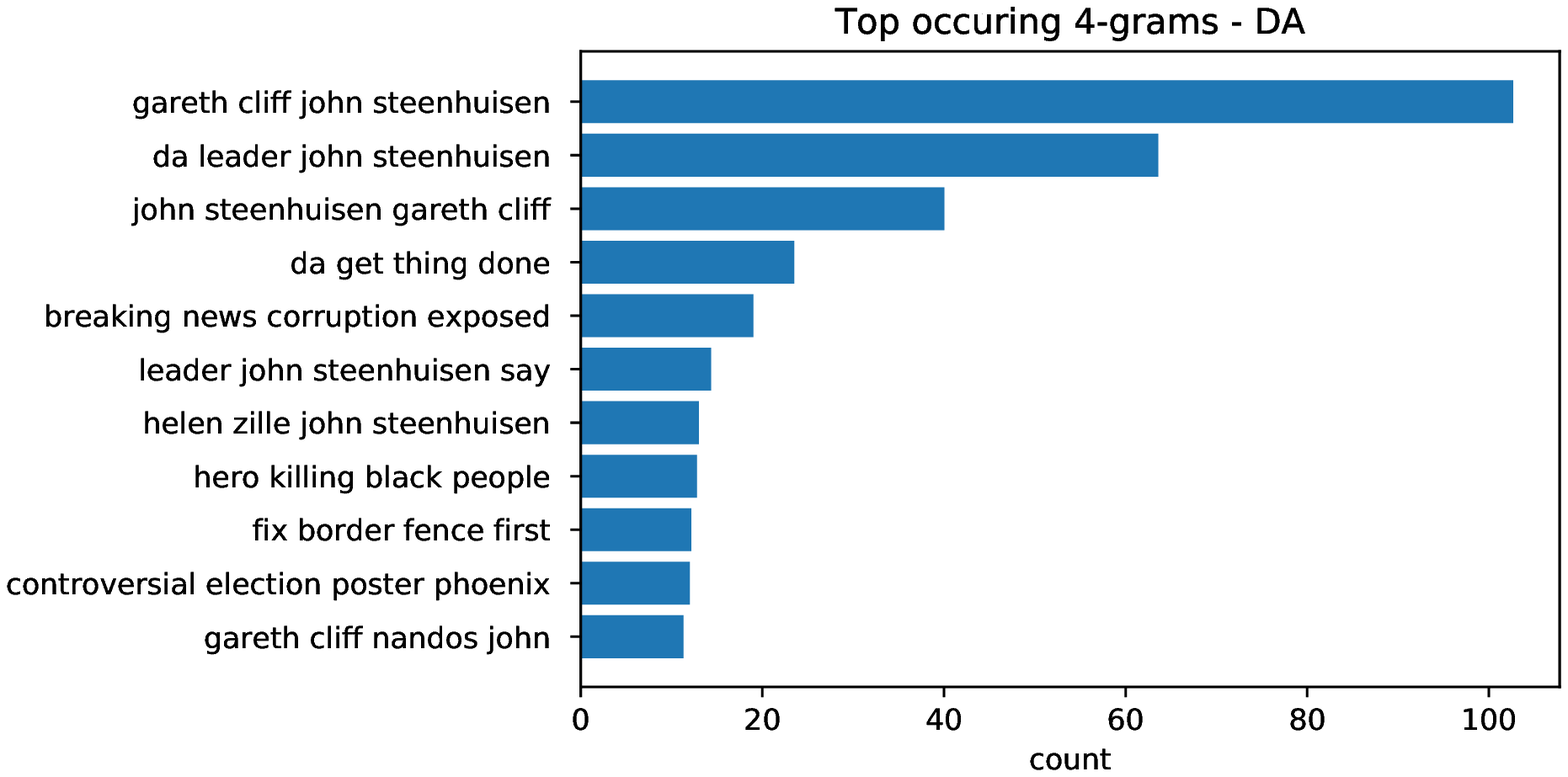}}
    \caption{DA negative sentiment topics word cloud (a),(b) and DA recurring themes using 4-grams (c)}
\end{figure}

\paragraph{EFF}Concerns associated with the EFF were issues around the looting of VBS with requests from users for the party to pay back the money. Expropriation of land has always been one of EFF’s main policy agenda and continues to be a major topic in the party as can be observed in the most frequent 4-gram associated with negative sentiment tweets for the EFF. Another recurring topic under the EFF is on their open-border policy. A number of South African patriots have expressed concerns about the policy and sent cautionary warnings of withholding their votes due to the party's stance on illegal immigration after the party's president mentioned at an EFF Presser that neighbouring countries should
“find creative ways” to enter South Africa.

\begin{figure}[htp]
    \centering
    \subfigure[]{\includegraphics[width=0.2\linewidth, height = 3.5cm]{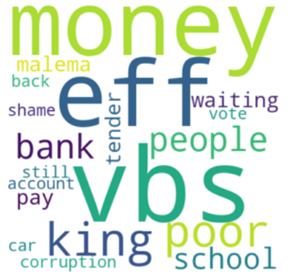}}
    \subfigure[]{\includegraphics[width=0.2\linewidth, height = 3.5cm]{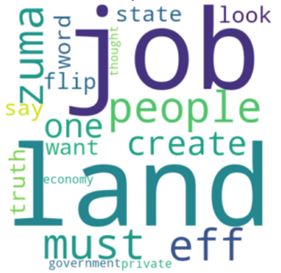}}
    \subfigure[]{\includegraphics[width=0.5\linewidth, height = 3.5cm]{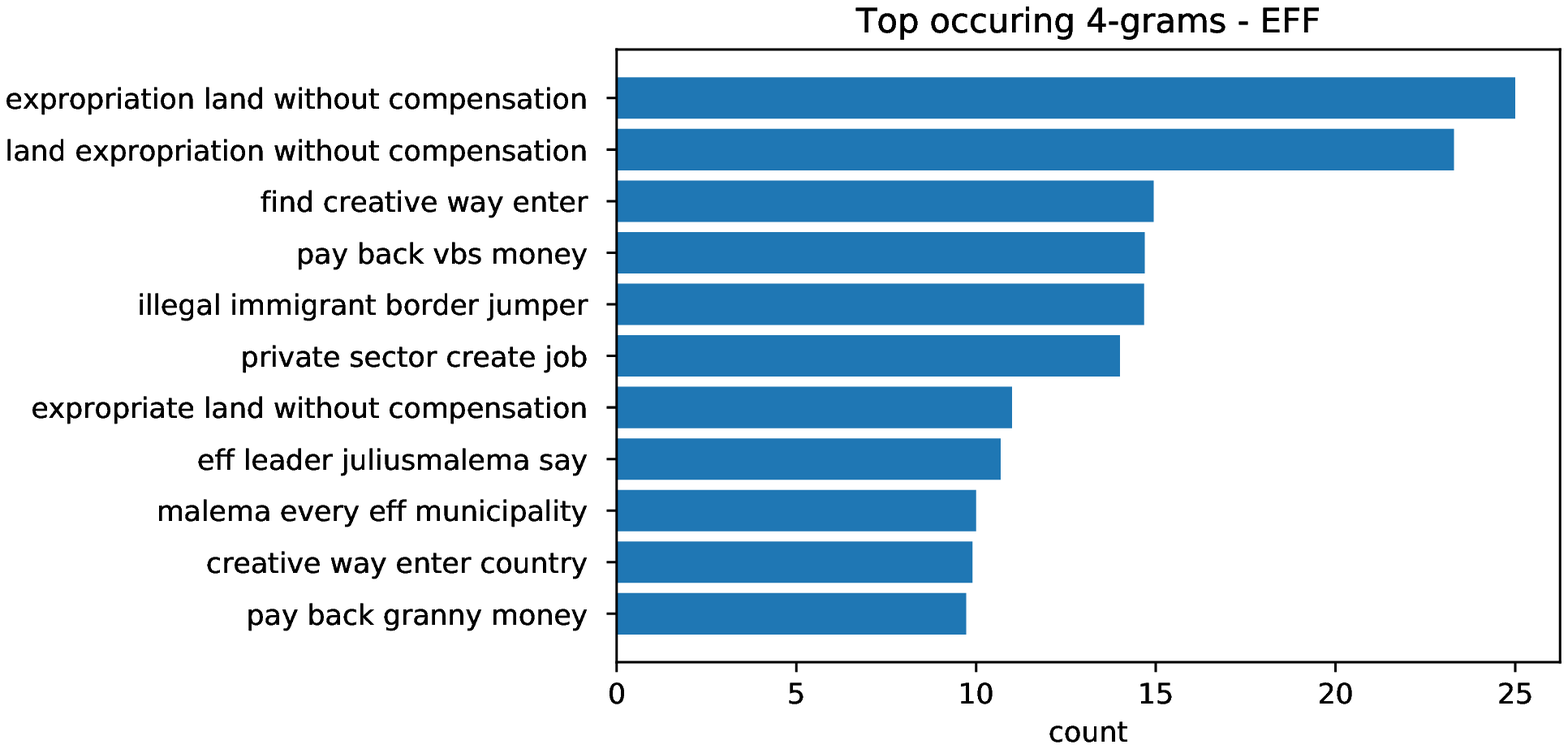}}
    \caption{EFF negative sentiment topics word cloud (a),(b) and EFF recurring themes using 4-grams (c)}
\end{figure}

\paragraph{ACTIONSA}The main concern in the negative sentiment tweets expressed towards the newly formed party is centered around their stance on illegal immigration, which has been viewed by many as draconian and xenophobic. Another topic of concern expressed in negative sentiment tweets concerned omission of the party’s name on the ballot paper for the elections. ActionSA’s 4-grams covers themes immigration and on expropriation of land - a topic commonly associated with the EFF. The latter is due to expressed concerns of ActionSA’s lack of planning and public interest when addressing topics related to land reform in South Africa.

\begin{figure}[htp]
    \centering
    \subfigure[]{\includegraphics[width=0.2\linewidth, height = 3.5cm]{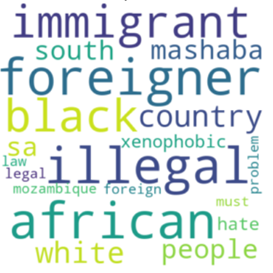}}
    \subfigure[]{\includegraphics[width=0.2\linewidth, height = 3.5cm]{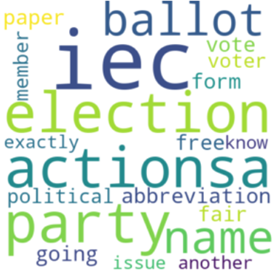}}
    \subfigure[]{\includegraphics[width=0.5\linewidth, height = 3.5cm]{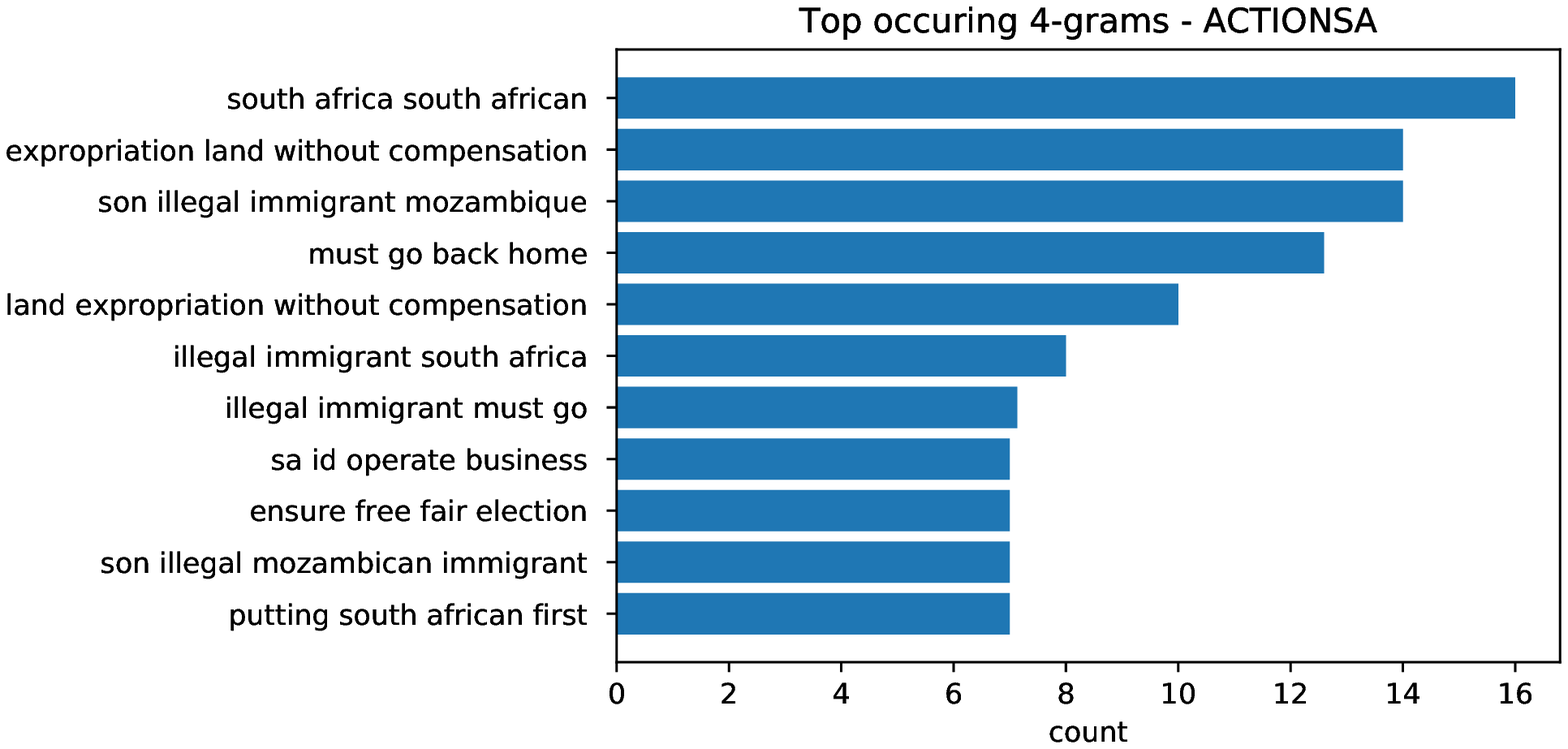}}
    \caption{ACTIONSA negative sentiment topics word cloud (a),(b) and ACTIONSA recurring themes using 4-grams (c)}
\end{figure}

\section{CONCLUSION AND DISCUSSION}

This study illustrated how social media sentiment analysis and topic modelling was used to understand the opinions shared by four different political parties in the South African context. Furthermore, based on this analysis, context into the positive and negative sentiments were highlighted. To predict political sentiment, Twitter data was processed and a small subset of the data manually labelled for positive and negative sentiment using a lexicon dictionary as a guide. A semi-supervised approach was followed to predict political sentiment using the graph-based method, label propagation, to learn from and propagate the small manually labelled sentiment to the unlabelled data. This method was used to address the gap in the current under-resourced sentiment models for the South African political landscape. The label propagation model using Skip-gram vector representation resulted in the best model performance across different metrics compared to TF-IDF and CBOW representations. This model was used to label the sentiment examples for the four political parties. The sentiment scores in this study indicated that the current ruling party (ANC) had the greatest negative score, with ActionSA constituting most of the positive sentiments as compared to any of the other political parties in this study. Topic modelling was used to extract topics uncovering the main concerns associated with each of the party’s negative tweets with the main concerns towards the ruling party being centred around corruption, incompetency, and Eskom. 

The appointment of Cyril Ramaphosa came with a promise to deal with the country’s insuperable corruption and three years later, corruption is still a major concern for the South African public as seen through topic modelling. The negative sentiment expressed towards the ruling party does not come as a surprise given these are only a few of the many recurring issues that the ruling party is having challenges addressing. Most of these issues are mainly attributed to incompetency, leading to failure to address issues of employment, service delivery, crime and issues faced by majority of the poor in the country. Internal factional battles within the party have also not made it easy
for the party to deal with critical issues of the country. 

The last research question was whether sentiment towards political parties drives voting intentions. Some studies have shown the predictability of sentiment analysis for election turnout \cite{jain_kumar_2017, oikonomou_2018, budiharto_meiliana_2018}. For our local government elections, this question can be answered by referencing the voting results as an empirical study would require additional information to be collected to extensively answer. Given the observed high negative opinions expressed towards the ruling party, it can be assumed that this sentiment and the concerns it encapsulates are reflected by the public's actions at the ballot stations. The ANC has recorded its worst voting results with an all-time electoral low below 50\%. The ruling party managed to maintain votes in the poor regions of the country such as regions within the Eastern Cape but
suffered major losses in regions where they previously had a stronghold such as Soweto and eThekwini. ActionSA unseated the ruling party in some of its major voting districts. For a party that is less than a year old, there has to be positive contributing factors that the public resonates with to give it such a huge support.

The elections saw the worst voter turnout post-apartheid with youth voter turnout being the worst and the youth are majority who have access to social media platforms and vast online information capable of making informed voting decisions. The low voter turn-out is a troubling concern which, the ruling party’s Deputy Secretary General, attributed to failures of the ANC in a statement on the local government elections \cite{statement_2021}: \textit{“it is in the main an unambiguous signal to the ANC from the electorate. The low voter turnout, especially in traditional ANC strongholds, communicates a clear message:
The people are disappointed in the ANC with the slow progress in fixing local government, in ensuring quality and
consistent basic services, in tackling corruption and greed.”} 
\section{LIMITATIONS AND FUTURE WORK}

This study gave us a good sense of the political climate for local government elections however, the use of social media to determine political sentiment excludes a great number of South Africans without access to technology or Twitter.

The dataset that was used for sentiment predictions of the respective parties disregarded multi-sentiment tweets where one tweet contains expressed sentiment about more than one political party. This reduced the data associated with each of the political parties. Proposed future work to address this would be using an automatic approach such as target-dependent sentiment analysis tasks which can handle different sentiment projected towards different subjects in the same tweet.

Another limitation in the study is that only positive and negative sentiment were used which means the model is not trained to identify tweets that may not be expressing neither positive nor negative sentiment.

Due to time and cost constraints, we did the manual annotation of the tweets ourselves. Even though they were verified by
a political commentator, we are not experts in political linguistics.

Based on \cite{Zhu2002LearningFL} and literature \cite{tai_kao_2013,10.5555/1609067.1609142, yang_shafiq_2018, rajadesingan_liu_2014} presenting the success of Label Propagation, it was the only semi-supervised approach that was explored in the study. As part of future proposed work, it would be good to also test different ways of propagating labels onto unlabelled data and doing a comparative study such as using other graph-based methods or wrapper-based methods like self and co-training approaches. 

To further improve model performance, we could also investigate a ‘transfer-learning’ approach by continuing the training of the powerful pre-trained Google models on our elections dataset to create an even better vector representation and also using these for topic modelling.

\section{Acknowledgements}

We would also like to thank the Data Science for Social Impact research group at the University of Pretoria who assisted in proof-reading the final submission. We would like to acknowledge the following funders: ABSA (who sponsor the UP ABSA Data Science Chair) and the National Research Foundation, South Africa.

\bibliographystyle{ACM-Reference-Format}
\bibliography{sample-base}

\end{document}